\documentclass[10pt,twocolumn,letterpaper]{article}

\usepackage{iccv}
\usepackage{times}
\usepackage{epsfig}
\usepackage{graphicx}
\usepackage{amsmath}
\usepackage{amssymb}
\usepackage{multirow}
\usepackage{caption} 
\usepackage{mathtools}
\DeclarePairedDelimiter\ceil{\lceil}{\rceil}
\DeclarePairedDelimiter\floor{\lfloor}{\rfloor}
\usepackage{transparent}
\usepackage[accsupp]{axessibility}

\usepackage[pagebackref=true,breaklinks=true, colorlinks,bookmarks=false]{hyperref}

\iccvfinalcopy % *** Uncomment this line for the final submission

\def\iccvPaperID{4612} % *** Enter the ICCV Paper ID here

\ificcvfinal\fi

\begin{document}

%%%%%%%%% TITLE
\title{Multiscale Representation for Real-Time Anti-Aliasing Neural Rendering}

% \author{Dongting Hu$^{1}$ \quad Zhenkai Zhang$^{1}$ \quad Tingbo Hou$^{2}$ \quad Tongliang Liu$^{3}$ \\ \quad Huan Fu$^{4,*}$ \quad Mingming Gong$^{1,}$\thanks{Equal contribution} \vspace{0.3em} \\
% {\normalsize $^1$University of Melbourne} \quad
% {\normalsize $^2$Google} \quad 
% {\normalsize $^3$University of Sydney} \quad
% {\normalsize $^4$Alibaba Group}
% }

\author{Dongting Hu \\
University of Melbourne\\
\and
Zhenkai Zhang\\
University of Melbourne\\
\and
Tingbo Hou\\
Google \\
\and
Tongliang Liu\\
University of Sydney\\
\and
Huan Fu$^{*}$\\
Alibaba Group\\
\and
Mingming Gong\thanks{Equal contribution}\\
University of Melbourne\\
}

\maketitle
% Remove page # from the first page of camera-ready.
\ificcvfinal\thispagestyle{empty}\fi

%%%%%%%%% ABSTRACT
\begin{abstract}
The rendering scheme in neural radiance field (NeRF) is effective in rendering a pixel by casting a ray into the scene. However, NeRF yields blurred rendering results when the training images are captured at non-uniform scales, and produces aliasing artifacts if the test images are taken in distant views. 
To address this issue, Mip-NeRF proposes a multiscale representation as a conical frustum to encode scale information. Nevertheless, this approach is only suitable for offline rendering since it relies on integrated positional encoding (IPE) to query a multilayer perceptron (MLP). 
To overcome this limitation, we propose mip voxel grids (Mip-VoG), an explicit multiscale representation with a deferred architecture for real-time anti-aliasing rendering. Our approach includes a density Mip-VoG for scene geometry and a feature Mip-VoG with a small MLP for view-dependent color. Mip-VoG represents scene scale using the level of detail (LOD) derived from ray differentials and uses quadrilinear interpolation to map a queried 3D location to its features and density from two neighboring down-sampled voxel grids.
To our knowledge, our approach is the first to offer multiscale training and real-time anti-aliasing rendering simultaneously. We conducted experiments on multiscale dataset, results show that our approach outperforms state-of-the-art real-time rendering baselines.

\end{abstract}

%%%%%%%%% BODY TEXT
\section{Introduction}
\label{sec:intro}
The realm of computer vision and graphics is marked by the captivating yet formidable challenge of novel view synthesis. In recent times, neural volumetric representations, most notably the neural radiance field (NeRF)~\cite{mildenhall2021nerf}, have emerged as a promising breakthrough in reconstructing intricate 3D scenes from multi-view image collections. 
NeRF employs a coordinate-based multilayer perceptron (MLP) architecture to map a 5D input coordinate (including 3D spatial position and 2D viewing direction) to intrinsic scene attributes (namely, volume density and view-dependent emitted radiance) at that precise location. The pixel rendering process in NeRF involves casting a ray through the pixel into the scene, extracting the scene representation for points sampled along the ray, and ultimately fusing these components to produce the final color output.

While this rendering methodology excels when the training and testing images share a uniform resolution, challenges arise when the training images encompass varying resolutions. This discrepancy in resolutions leads NeRF to produce blurred rendering outputs due to the altered pixel footprints originating from diverse scales. Besides, in case that test viewpoints significantly deviate from the spatial distance of the training views, the sample rate for per pixel would be inadequacy, thereby results in aliasing artifacts.

\begin{table}
% \tiny
\centering
\resizebox{.45\textwidth}{!}{%
    \begin{tabular}{l|c|c|c}
        \hline
        \multirow{2}{*}{Method} & 
        \multicolumn{1}{|p{1.7cm}|}{\centering Multiscale\\Training} &
        \multicolumn{1}{|p{1.7cm}|}{\centering Real-time\\Rendering} & 
        \multicolumn{1}{|p{1.9cm}}{\centering Anti-aliasing\\Rendering} \\
        \hline \hline
         Mip-NeRF~\cite{barron2021mip} & \checkmark &  & \checkmark\\
         SNeRG~\cite{hedman2021baking}  &    & \checkmark&  \\
         MobileNeRF~\cite{chen2022mobilenerf}  &  & \checkmark& \checkmark\\ 
         Ours  & \checkmark & \checkmark & \checkmark \\
        \hline
    \end{tabular}
    }
\caption{Our method is the first one concurrently addresses multiscale training, real-time and anti-aliasing rending.}
\label{tab:intro}
\end{table}

To surmount this challenge, Mip-NeRF~\cite{barron2021mip} emerges as a noteworthy solution, presenting a continuously-valued scale representation for coordinate-based models. Mip-NeRF introduces a pioneering technique, known as integrated positional encoding (IPE), which facilitate the scene representation with the knowledge of scale.
Departing from the conventional ray-based approach, Mip-NeRF adopts a novel rendering strategy involving conical frustums. This representation not only enables effective multiscale training but also tangibly mitigates the persisting issue of aliasing artifacts. 
However, it's important to note that this approach relies on querying the network with the scale-variant IPE. As a consequence, the integration of pre-cache techniques that circumvent positional encoding~\cite{hedman2021baking}, remains elusive within the current Mip-NeRF's formulation.

This paper introduces a novel multiscale representation, termed "Mip-VoG" (Mip Voxel Grids, drawing inspiration from "mipmap"), which addresses the challenge of training on images with varying scales and enables real-time anti-aliasing rendering during the inference stage (Tab.~\ref{tab:intro}). Our approach commences by unveiling a "deferred" NeRF variant, wherein both scene geometry and color attributes are explicitly stored within the Mip-VoG framework. The intricate task of decoding view-dependent effects is executed using a compact multilayer perceptron (MLP) that efficiently processes each pixel only once.
Instead of pre-training and baking a continuous NeRF into grids~\cite{hedman2021baking}, we treat voxel values as parameters and directly optimize them as seen in the work by~\cite{fridovich2022plenoxels, sun2022direct}. 
Despite the ``mip'' nomenclature, Mip-VoG only maintains one density voxel grid and one feature voxel grid that represent the high-frequency spatial attributes. This structure permits the inference of lower frequency representations through a progressive down-sampling process, achieved via low-pass filters and interpolation algorithms~\cite{ewins1998mip}. 
Given a single 3D point sampled from a camera ray, Mip-VoG intelligently determines the level of detail (LOD) via ray differentials~\cite{igehy1999tracing}. The LOD calculation is pivotal, as it establishes a pixel-to-voxel ratio that represents the point's footprint on the full-resolution voxel grids. 
As a consequence, scene properties corresponding to this point are sampled by interpolating between two adjacent down-sampled level grids. Notably, camera rays cast from a low-resolution frame are spatially represented over a wider area, yielding a higher sample rate and capturing more lower-frequency information.
During inference phrase, we pre-compute the voxel grids at each integer level and  subsequently convert them to the sparse voxel grid data structure used by SNeRG~\cite{hedman2021baking} to increase the rendering speed.

% We evaluate our method on the widely studied NeRF datasets such as Synthetic-NeRF~\cite{mildenhall2021nerf} and Multiscale-NeRF~\cite{barron2021mip} following MipNeRF's multi-scale setting~\cite{barron2021mip}.
% The results show that, compared to state-of-the-art real-time techniques, our multiscale representation successfully tackled multiscale training and preserve both low and high frequencies from the multiscale detest. The result also show that, based on proposed representation, our methods achieve high accuracy in anti-aliasing rendering.

We conducted a comprehensive evaluation of our approach using well-established NeRF datasets, including Synthetic-NeRF~\cite{mildenhall2021nerf} and Multiscale-NeRF~\cite{barron2021mip}, in line with the multi-scale framework outlined in Mip-NeRF~\cite{barron2021mip}. Our findings underscore the validity of our multiscale representation in effectively addressing complex multiscale training scenarios, while successfully preserving both low and high-frequency components inherent in the multiscale dataset. Comparative analysis against state-of-the-art real-time techniques reveals that our proposed approach excels in mitigating multiscale challenges, yielding impressive results. Furthermore, our method proves instrumental in achieving remarkable accuracy in anti-aliasing rendering, bolstering its applicability and potential impact.
%------------------------------------------------------------------------
\section{Related Works}
\label{sec:related_work}
\paragraph{Scene Representation for View Synthesis}
Numerous scene representations have been proposed to tackle the intricate task of view synthesis. Approaches such as Light Field Representation~\cite{davis2012unstructured,levin2010linear,levoy1996light,shi2014light} and Lumigraph~\cite{gortler1996lumigraph, buehler2001unstructured} directly interpolate input images, albeit necessitating dense input data for novel view synthesis. In an effort to reduce the demand for exhaustive capture, subsequent studies represent light fields as neural networks~\cite{sitzmann2021light,attal2022learning}. Layered Depth Images~\cite{dhamo2019peeking,shade1998layered,shih20203d,tulsiani2018layer} alleviate the requirement of input denseness, but their effectiveness hinges on the accuracy of depth maps for rendering photo-realistic images. Recent advancements have introduced methods to estimate Multiplane Images (MPIs)\cite{flynn2019deepview,li2020crowdsampling,mildenhall2019local,srinivasan2019pushing, wizadwongsa2021nex, zhou2018stereo} for scenes with forward-facing viewpoints, and voxel grids\cite{sitzmann2019deepvoxels, lombardi2019neural} for inward-facing scenes. 
Mesh-based representations~\cite{dhamo2019peeking,shade1998layered,shih20203d,tulsiani2018layer,munkberg2022extracting,nicolet2021large, hasselgren2022shape} constitute another notable category within the view synthesis realm, offering real-time rendering potential through optimized rasterization pipelines. However, these methods necessitate template meshes as priors to overcome gradient-based optimization challenges.

Recently, NeRF~\cite{mildenhall2021nerf} emerges as a popular method for novel view synthesis. By using a MLP as an implicit and continuous volumetric representation, NeRF maps from a 3D coordinate to the volume density and view-dependent emission at that position. The success of NeRF brings numbers of attention into neural volumetric rendering for view synthesis.
Many follow-on works have extended NeRF to generative models~\cite{chan2021pi, trevithick2021grf, schwarz2020graf, niemeyer2021giraffe}, generalization~\cite{wang2021ibrnet, yu2021pixelnerf}
dynamic scenes~\cite{ost2021neural, li2021neural, pumarola2021d, martin2021nerf, rudnev2022nerf}, relighting\cite{bi2020neural, srinivasan2021nerv, zhang2021nerfactor}, and editing~\cite{park2021nerfies, park2021hypernerf, tretschk2021non, yuan2022nerf, liu2021editing, yang2022neumesh}, etc. 
Rendering an image via NeRF necessitates querying an extensive neural network at multiple 3D locations per pixel, resulting in approximately a minute per frame rendering time. Recent advancements seek to enhance NeRF's rendering efficiency through explicit representation leveraging~\cite{chen2022tensorf,sun2022direct,fridovich2022plenoxels, wizadwongsa2021nex}, or by segmenting the scene into sub-regions with smaller neural networks~\cite{reiser2021kilonerf, rebain2021derf}. 

% DVGO~\cite{sun2022direct} directly optimizes the voxel grids by gradient-descent, but still keep a small MLP to predict the RGB color of 3D location. Plenoxel~\cite{fridovich2022plenoxels} directly optimizes the voxel grid without any involvement of neural networks at all. 
% DVGO \cite{sun2022direct} uses the simplest dense grid data structure in fully Pytorch implementation. Plenoxels \cite{fridovich2022plenoxels} model the coefficients of spherical harmonic for view-dependent colors and realize a fully explicit (without MLP) representation. 
% Instant-NGP \cite{muller2022instant} uses hash-table and hybrid representations for both densities and colors. Instant-NGP further improves the training time using C/C++ and fully-fused CUDA implementation. TensoRF \cite{chen2022tensorf} improves the memory footprint and scalability of the dense grid via tensor decomposition and directly modeling the low-rank components.KiloNeRF~\cite{reiser2021kilonerf} utilizes thousands of tiny MLPs to accelerate the rendering. FastNeRF~\cite{garbin2021fastnerf} re-factorize the NeRF and cache the scene interest for  efficient running. However, these methods need to train an implicit model (e.g., NeRF) from scratch before their final conversion to the certain representation, results in the heavy pre-train burden and lengthy optimization process.

\paragraph{Real-time Neural Rendering}
A series of researchers has emerged to address the imperative demand for real-time rendering capabilities.  PlenOctrees~\cite{yu2021plenoctrees} introduces an innovative spherical harmonic representation of radiance, seamlessly transitioning it into an octree data structure.  FastNeRF~\cite{garbin2021fastnerf} strategically restructures NeRF through refactoring, incorporating a dense voxel grid to efficiently cache the scene of interest for accelerated rendering. iNGP~\cite{muller2022instant} uses a hash-table to store the feature vectors and combining with fully-fused CUDA kernels to accelerate rendering processes. 
SNeRG~\cite{hedman2021baking} proposes a deferred architecture and extracts the scene properties from a pre-trained model into a sparse grid data structure. On a divergent trajectory, MobileNeRF~\cite{chen2022mobilenerf} adopts a unique approach, representing the scene using textured polygons and harnessing polygon rasterization to generate pixel-level features. These features, in turn, are decoded via a compact view-dependent MLP. Nevertheless, it's important to note that the explicit representations harnessed by these approaches lack scale-agnostic adaptability, thus the efficacy in learning from training images with multiple resolutions is limited. Our primary focus in experimental comparisons resides with SNeRG and MobileNeRF, given their established performance on resource-constrained devices without CUDA access.

\paragraph{Reducing Aliasing in Rendering}
One straightforward solution to mitigate aliasing for coordinate-based neural representations is supersampling~\cite{whitted2005improved}, which requires casting multiple rays through pixel during rendering to get the final result. While powerful in its anti-aliasing effects, supersampling exacerbates the already time-intensive rendering procedure of NeRF, thereby confining its utility primarily to offline rendering scenarios. To improve the efficiency, Mip-NeRF~\cite{barron2021mip} proposes to cast a conical frustum into the scene space and render the 3D region instead of a single point. This approach avoids heavy computation burden by approximating the 3D region rendering using gaussian, their algorithm queries the network by IPE of the 3D input region to output the final density and radiance. Due to the heavy reliance on the network to decode the scale information, Mip-NeRF cannot leverage pre-cache techniques to enable real-time capability.
Another common technique for reducing aliasing is pre-filtering~\cite{olano2010lean, kaplanyan2016filtering, wu2019accurate, bruneton2011survey}, which pre-filters the maps on a coarse mesh \eg color maps, normal maps, linearly and separately. This strategy involves the pre-filtration of various maps on a coarse mesh, such as color maps and normal maps, independently and linearly. Notably, pre-filtering transfers the computational load to a pre-rendering stage, rendering it well-suited for real-time rendering scenarios. A widely adopted method in 3D rendering applications is mip mapping~\cite{ewins1998mip, williams1983pyramidal, crassin2009beyond}. A serialization of images or textures, each of which is a progressively lower resolution representation of the previous one, are pre-computed ahead of time to increase rendering speed and reduce aliasing artifacts for real-time inference. Traditionally, mip mapping is integral to the texture mapping process for 3D meshes. Expanding upon this concept, we extend mip mapping to the realm of 3D neural volumetric rendering. Our approach involves applying mip mapping to the voxel grids data structure, rather than a mere 2D map. This extension capitalizes on the advantages of mip mapping to boost rendering efficiency and reduce aliasing artifacts within the domain of neural volumetric rendering.

% The natural scenes exhibit smoothness, thus motivating the use of a multi-resolution decomposition.
% NSVF~\cite{liu2020neural} and several concurrent works~\cite{hedman2021baking} adopt a multistage, coarse to fine strategy in which regions of the feature grid are progressively refined and culled away as necessary.  Octrees, which are a tree-based data structure, are often used in 3D graphics and 3D game engines, and it's also used in signed distance~\cite{zeng2013octree,vespa2019adaptive}. It removes the empty from the space without color by the function of pruning and subdivides the remaining grid to get an accurate enough shape to achieve the effect of reducing blending, which is also used in Plenoxel~\cite{fridovich2022plenoxels}, which named as PlenOctree.

\paragraph{Relation to DVGO, iNGP and ZipNeRF} There are some remarkable concurrent works study the voxel representation for efficient rendering. Still, there are some difference between Mip-VoG and these works. DVGO~\cite{sun2022direct} progressively optimize a higher resolution voxel grid in the training for finer details, but it does not considering a multi-scale representation. Besides, DVGO stores the implicit feature for radiance emission and predict the final pixel color by a shallow MLP. By contrast, our Mip-VoG stores the explicit value for diffuse color and implicit feature for view-dependent specular radiance.
iNGP~\cite{muller2022instant} introduces a hash encoding approach based on a multi-resolution structure for speedy high-quality image synthesis. Their approach simultaneously trains several dense grid with different scales and concatenates the feature from each for further prediction. This representation only works for single-scale images since its scale-invariant. In contrast, our method only retains a single voxel grid during training and can progressively sample from Mip-VoG with different LOD. A concurrent work, ZipNeRF~\cite{barron2023zipnerf} operates within a similar setting, where they integrate iNGP's grid pyramid using multi-sampling within the Mip-NeRF framework, whereas our work takes a distinct approach by incorporating the fundamental concept of ``mip'' directly into voxel grids.

%------------------------------------------------------------------------
% \section{Preliminaries}
% \input{Mip-VoG/sections/preliminary}
%------------------------------------------------------------------------
\section{Method}
\begin{figure*}[ht]
  \centering
    \centering
    \includegraphics[width=1.0\textwidth]{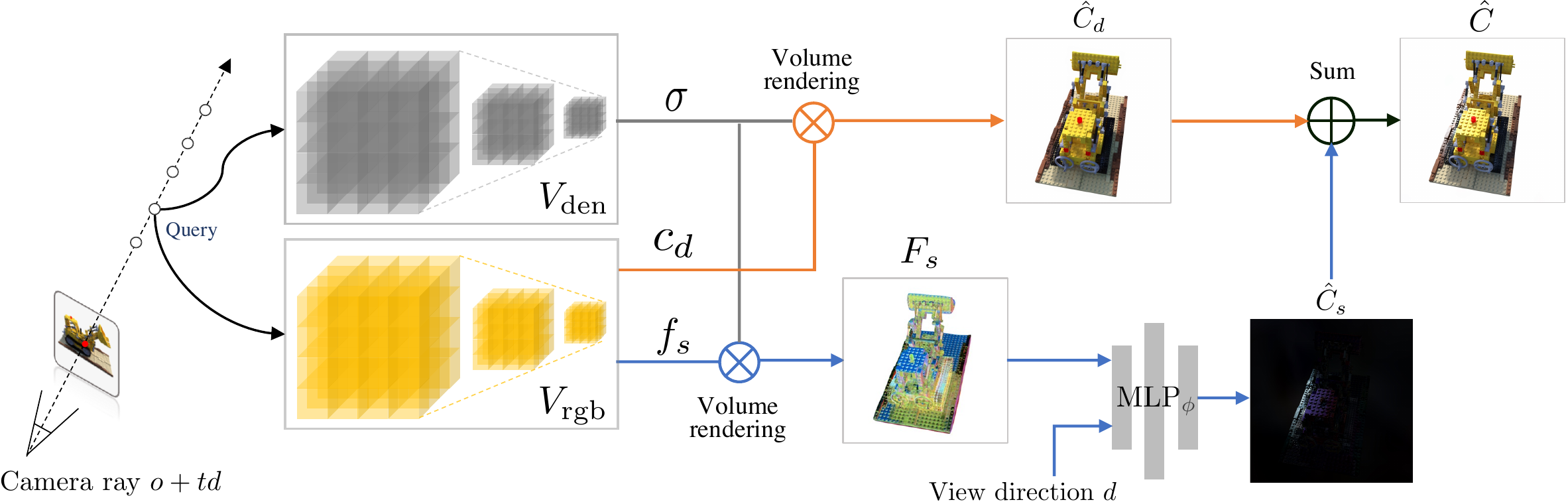}
  \caption{\textbf{Rendering framework overview.} We based our pipeline design on deferred NeRF~\cite{hedman2021baking} with explicit training of Mip-VoG. Given a point sampled from a camera ray, we query the density Mip-VoG ($V_{\text{den}}$) and color Mip-VoG ($V_{\text{rgb}}$) to predict the 1D volume density $\sigma$, 3D diffuse color $c_d$ and 4D feature vector $f_s$. We then aggregate the diffuse colors and feature vectors along the ray through volume rendering integral (Eq.~\ref{eqcd} and \ref{eqfs}), resulting in $\hat{C}_d$ and $F_s$. After that, a tiny MLP is used to predict a pixel-wise view-dependent specular color $\hat{C}_s$ by using the accumulated feature-vector $F_s$ together with the view direction $d$ (Eq.~\ref{eqcs}). The final color prediction $\hat{C}$ is the summation of diffuse color $\hat{C}_d$ and specular color $\hat{C}_s$ (Eq.~\ref{eqc}).}
    \label{fig:arch}
\end{figure*}

\subsection{Review of NeRF}
NeRF~\cite{mildenhall2021nerf} uses a ${\mathrm{MLP}}$ parameterized by $\theta$ as a continuous volumetric function to represent a scene. The network takes input as the view direction $d$ and 3D coordinate $r(t)$ sampled from a camera ray $r(t)=o+td$, and predict the volume density $\sigma$ at that 3D position together with the view-dependent radiance $c$ from that view direction:
\begin{equation}
\sigma(t), c(t) = \mathrm{MLP}_\theta(r(t), d).
\end{equation}
A vital assumption made by NeRF is to model the density $\sigma$ only depend on location, while emitted color is conditional on 3D coordinate $r(t)$ and view direction $d$.
In the rendering procedure, NeRF takes the predicted densities and emissions $\{\sigma(t_i), c(t_i)\}_{i=1}^N$ along the ray casted from a pixel, and approximate a volume rendering integral~\cite{max1995optical} to derive the final color of that pixel:
\begin{equation}
\begin{split}
\hat{C}(r) =& \sum_{i=1}^{N}T_i\Big(1-\exp\big(-\sigma(t_i)\delta(t_i)\big)\Big)c(t_i), \\
\text{with } T_i &= \exp(-\sum_{j=1}^{i-1}\sigma(t_j)\delta_j),
\end{split}
\end{equation}
where $\delta(t_i)=t_{i+1}-t_{i}$ is the distance between adjacent samples. One can find that rendering a single ray for each pixel requires evaluating the MLP hundreds of times, resulting in significantly slow rendering speed.

\subsection{Review of Deferred NeRF}
As discussed previously, real-time rendering can be achieved by pre-computing as many as scene properties. While the scene geometry (volume density) can be directly stored, NeRF relies on a continuous function to represent view-dependent effects. To address this problem, SNeRG~\cite{hedman2021baking} introduces a residual architecture that first caches the point-wise pre-trained volume density $\sigma(t)$, diffuse color $c_d(t)$ and 4-dimension feature vector $f_s(t)$. The pixel-wise diffuse color and feature vector are obtained through volume rendering (same as NeRF):
% \begin{equation}
% \hat{C}_d(r) = \sum_{i=1}^{N}T_i\Big(1-\exp\big(-\sigma(t_i)\delta(t_i)\big)\Big)c_d(t_i),\\
% \label{eqcd}
% \end{equation} 
% \begin{equation}
% F_s(r) =
% \sum_{i=1}^{N}T_i\Big(1-\exp\big(-\sigma(t_i)\delta(t_i)\big)\Big)f_s(t_i).\\
% \label{eqfs}
% \end{equation}
\begin{align}
    \label{eqcd}
    \hat{C}_d(r) &= \sum_{i=1}^{N}T_i\Big(1-\exp\big(-\sigma(t_i)\delta(t_i)\big)\Big)c_d(t_i),\\
    \label{eqfs}
    F_s(r) &= \sum_{i=1}^{N}T_i\Big(1-\exp\big(-\sigma(t_i)\delta(t_i)\big)\Big)f_s(t_i).
\end{align}
Then, a tiny MLP parameterized by $\phi$, which forwards once for each pixel based on the feature $F_s(r)$ and view direction $d$, is applied to predict the pixel-wise specular color as a view-dependent residual:
\begin{equation}
\hat{C}_s(r) = {\textrm{MLP}}_\phi(F_s(r), d).
\label{eqcs}
\end{equation}
The final color of the pixel is obtained by the summation of diffuse color and specular color:
\begin{equation}
\hat{C}(r) = \hat{C}_d(r) + \hat{C}_s(r).
\label{eqc}
\end{equation}
Vanilla SNeRG involves pre-training a continuous representation first and caching voxel-wise $\sigma$, $c_d$ and $f_s$ into a sparse voxel grid. In contrast, our method directly learns an explicit representation from scratch, which can be directly used for efficient multiscale representation. We optimize $\sigma \in \mathbb{R}$ in one density Mip-VoG $V_{\text{den}}$ and $c_d \in \mathbb{R}^3$ together with $f_s \in \mathbb{R}^4$ in one color Mip-VoG $V_{\text{rgb}}$. The overview of our framework is shown in Fig. \ref{fig:arch}. In the following section, we present the query algorithms of Mip-VoG. 

% \begin{figure}[h!]
%     \centering
%     \begin{minipage}{\linewidth}
% \centerline{\includegraphics[width=\textwidth]{iccv2023AuthorKit/figures/method_arch.pdf}}
%     \end{minipage}
%     \caption{Illustration of architecture}
%     \label{fig:arch}
% \end{figure}

\subsection{Mip Voxel Grids}
Mip-VoG harnesses the potential of a sequence of progressively down-sampled "much in little" voxel grids to facilitate scene queries at specific 3D coordinates. When sampling from Mip-VoG for a singular point within the spatial domain, a pivotal initial step involves the calculation of Level of Detail (LOD), which serves as a representative ``correct" scale with regard to the complete-resolution voxel grids (section~\ref{sec_lod}). Subsequent procedures encompass filtering and down-sampling operations applied to the comprehensive voxel grids, resulting in the generation of voxel grids at progressively lower scales. The value sampled from the Mip-VoG, corresponding to the specific LOD, is then obtained through interpolation, which bridges the information across two distinct scales of voxel grids at neighboring levels (section~\ref{sec_sampling}).

To streamline notation, we simplify the representation by omitting subscripts, condensing $V_{den}$ and $V_{rgb}$ into a singular notation: $V^{(0)}$ signifies the original voxel grids at level 0. By analogy, the terms $V^{(1)}$, $V^{(2)}$, $V^{(k)}$ denote the voxel grids after down-sampling, signifying levels 1, 2, and $k$ respectively.

\subsubsection{Level of Detail}
\label{sec_lod}
\begin{figure}[h!]
    \centering
    \begin{minipage}{\linewidth}
\centerline{\includegraphics[width=\textwidth]{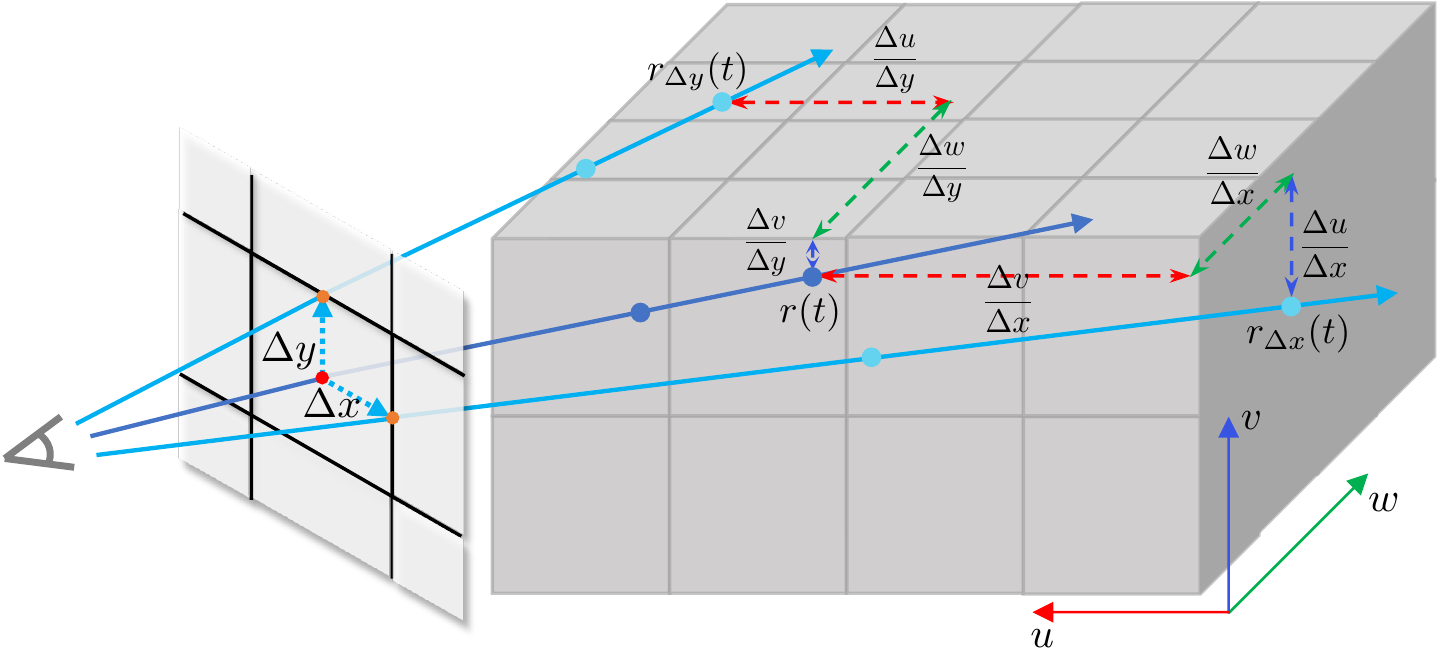}}
    \end{minipage}
    \caption{\textbf{Demonstration of ray differentials.} To determine the LOD of a point on the camera ray, we first cast two distinct rays, each generated by introducing an offset equivalent to half the pixel size along the x and y axes within the screen space. Subsequently, for every point situated along the original ray trajectory $r(t)$, we compute unit distance between this point its neighbours $r_{\Delta x}(t)$, $r_{\Delta y}(t)$ along three axis on the full-resolution voxel grids $V^{(0)}$, as $\Delta u$, $\Delta v$, $\Delta w$. The LOD is finally calculated based on the largest length of the voxel-to-pixel ratio (Eq.~\ref{eq:lod}).}
    \label{fig:raydiff}
\end{figure}

% We use the idea of ray differential to derive the level of detail (LOD). 
Given a single ray cast through a pixel, ray differentials represent a pair of differentially offset rays slightly above or to the right of the original ray~\cite{igehy1999tracing, akenine2019texture}. We extend this idea to the volumetric rendering with voxel grids.
Denote $u, v, w$ as the unit coordinates as regard to the voxel space $V^{(0)}$, for a pixel of the frame with image space coordinates $x$ and $y$, the ray differentials are defined as derivatives of the ray's footprint on voxel space with respect to image space coordinates ($x$ and $y$):
\begin{equation}
% \begin{split}
\frac{\partial V^{(0)}}{\partial x } = 
\Big\langle  
\frac{\partial u}{\partial x }\  
\frac{\partial v}{\partial x }\
\frac{\partial w}{\partial x }
\Big\rangle	
, ~~
\frac{\partial V^{(0)}}{\partial y } = 
\Big\langle  
\frac{\partial u}{\partial y }\
\frac{\partial v}{\partial y }\
\frac{\partial w}{\partial y }
\Big\rangle	.
% \end{split}
\end{equation}
By applying a first-order Taylor approximation, we can get an expression for the extent of a pixel’s footprint in voxel space based on the voxel-to-pixel spacing:
\begin{equation}
% \begin{split}
\frac{\partial V^{(0)}}{\partial x } \approx
\Big\langle  
\frac{\Delta u}{\Delta x }\  
\frac{\Delta v}{\Delta x }\
\frac{\Delta w}{\Delta x }
\Big\rangle	
, ~~
\frac{\partial V^{(0)}}{\partial y } \approx
\Big\langle  
\frac{\Delta u}{\Delta y }\
\frac{\Delta v}{\Delta y }\
\frac{\Delta w}{\Delta y }
\Big\rangle	.
% \end{split}
\end{equation}
Intuitively, this can be seen as the per-axis offset on the voxel grids made by the deviation of the ray in the image plane. As illustrated in Fig.~\ref{fig:raydiff}, we adopt $\Delta x, \Delta y$ as the half pixel size along the $x$ and $y$ axis of the image plane, since it can approximate the footprint of a pixel. Given the position of a point $r(t)$ and its neighbors $r_{\Delta x}(t), r_{\Delta y}(t)$ in the world space, the offset in voxel unit coordinate can be derived by the ratio between the distance per axis and the voxel size:
\begin{equation}
\Delta u = \frac{T_u}{V_u}, \
\Delta v = \frac{T_v}{V_v}, \
\Delta w = \frac{T_w}{V_w},
\end{equation}
where $T_u, T_v, T_w$ are per axis distance to the neighbors in the world space and $V_u, V_v, V_w$ are voxel size on $u, v, w$ along three axes. Then following the convention of mipmapping~\cite{heckbert1983texture, williams1983pyramidal}, the LOD $\lambda$ can be calculated as~\footnote{Since the number of voxels drops by 8x each level, we have $\log_8(\rho) = \log_2(\rho) / \log_2(8) = 1/3 * \log_2(\rho)$.}:
\begin{equation}
\begin{split}
  \lambda = 1/3 * \log_2(\rho) \\
\text{with }
\rho = \max \Biggl( \
% \begin{Bmatrix}
  & \sqrt{(\frac{\Delta u}{\Delta x})^2
  +(\frac{\Delta v}{\Delta x})^2
  +(\frac{\Delta w}{\Delta x})^2}\ , \\
  & \sqrt{(\frac{\Delta u}{\Delta y})^2
  +(\frac{\Delta v}{\Delta y})^2
  +(\frac{\Delta w}{\Delta y})^2}
  \ \Biggr).
% \end{Bmatrix}
\label{eq:lod}
\end{split}
\end{equation}

\subsubsection{Filtering and Sampling}
\label{sec_sampling}
\begin{figure}[h!]
    \centering
    \begin{minipage}{\linewidth}
    \centerline{\includegraphics[width=0.8\textwidth]{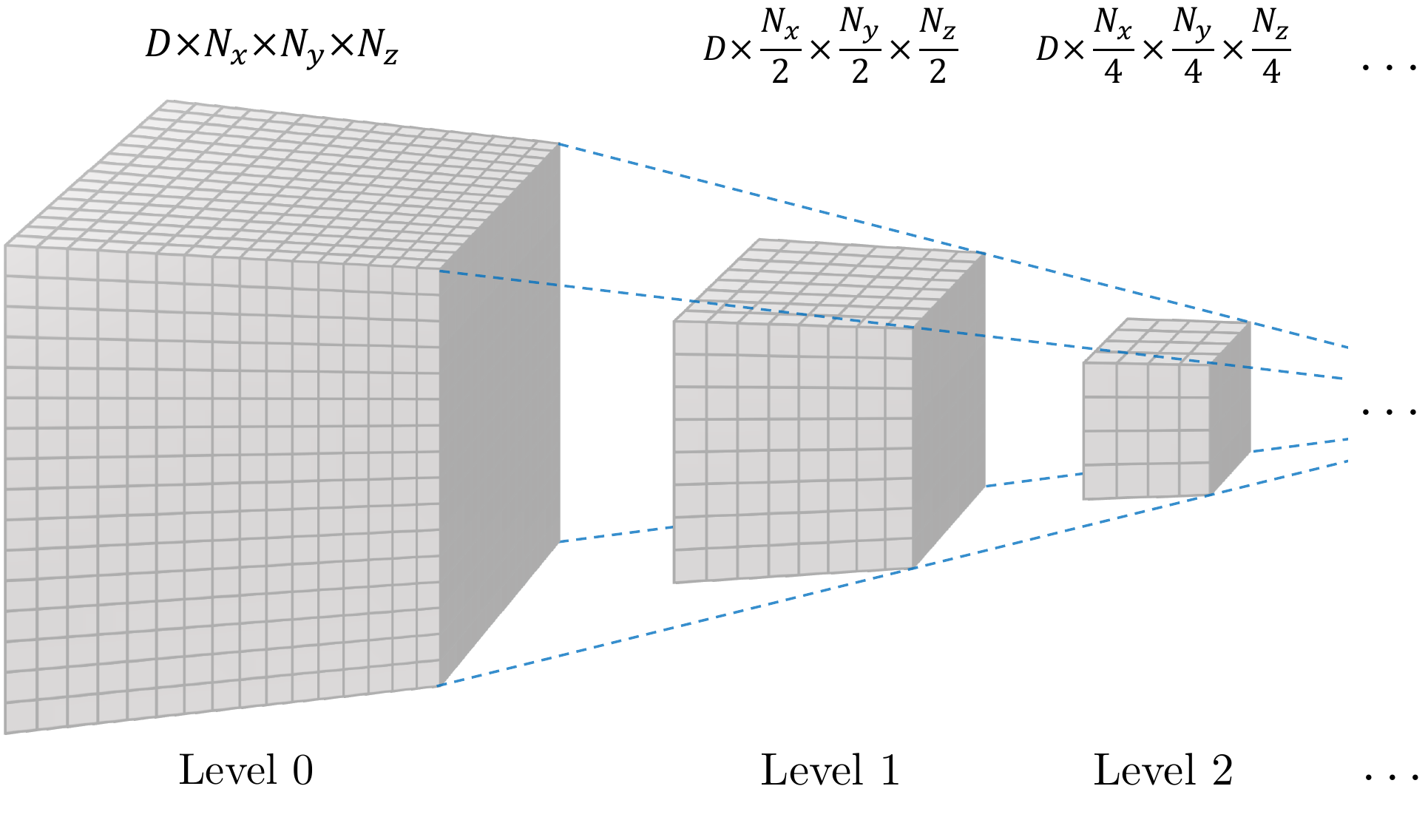}}
    \end{minipage}
    \caption{\textbf{Structure of Mip-VoG.} At each incremental increase in LOD, we first perform filtering on the previous voxel grids using low-pass filter, to remove high-frequency details. Subsequently, the filtered voxel grids are down-sampled to half of the previous resolution using linear interpolation. When querying the value for a 3D point within the space, Mip-VoG entails the interpolation of results derived from two adjacent voxel grids, selected in correspondence with the LOD of the point.}
    \label{fig:mipvog}
\end{figure}

Once LOD has been determined, the subsequent step entails sampling the relevant information from the voxel grid at the appropriate scale, as mandated by the Mip-VoG framework. To achieve "much in little" voxel grids, we progressively down-sample the original voxel grid $V^{(0)}$ into a series of successively lower resolutions. Prior to down-sampling, a pivotal preparatory step involves the application of a low-pass filter denoted as $\gamma$. This filter serves to mitigate high-frequency information, ensuring that the down-sampled representations are suitably refined and devoid of artifacts. Following most common mipmap techniques~\cite{lefebvre2005parallel}, the down-sampling process is conducted in a hierarchical fashion. With each increment in LOD by one level (integer), the default operation involves down-sampling the voxel grid using a scale of $1/2$:
\begin{equation}
    V^{(k+1)} = \ \Downarrow_{1/2}\big( \gamma(V^{(k)})\big),
\end{equation}
where $\Downarrow_{1/2}(.)$ represents the down-sampling with $1/2$ resolution. This process iterates, progressively generating new voxel grids at lower resolutions, thereby accommodating the spatial requirements of the specific LOD and maintaining the coherence of the Mip-VoG representation. In our approach, we employ linear interpolation as the method for down-sample filtering. To illustrate, consider the original voxel grid $V^{(0)}$ with dimensions ${D \times N_x \times N_y \times N_z}$. When generating $V^{(1)}$, the dimensions are adjusted through scaling to ${D \times N_x/2 \times N_y/2 \times N_z/2}$. The parameter $D$ signifies the dimension of the modality being considered. This process is extrapolated to achieve a feature voxel grid with higher LOD and correspondingly lower resolution. The resulting structure preserves the hierarchical nature of Mip-VoG and contributes to a coherent representation of the scene, as shown in Fig.~\ref{fig:mipvog}. 

To ensure the smooth continuity across non-integer LOD $\lambda$, we adopt quadrilinear interpolation to aggregate samples from two neighboring voxel grids at different levels (upper and lower). Let $f(V, p)$ denote the sampling function of the voxel grids, and $\mu \in \{\sigma, c_d, f_s\}$ represent the stores value, the interpolation process can be mathematically expressed as follows:
\begin{equation}
    \mu = (\ceil*{\lambda} - \lambda)f\big(V^{(\floor*{\lambda})},r(t)\big)
    +
    (\lambda - \floor*{\lambda})f\big(V^{(\ceil*{\lambda})},r(t)\big),
\end{equation}
where $\floor*{\cdot }$ and $\ceil*{\cdot }$ are the floor and ceiling function. This quadrilinear interpolation mechanism ensures that the sampled values are seamlessly blended between the two neighboring voxel grids, preserving the continuity of the representation across different LOD values.

\subsubsection{Optimization}
\label{sec_optim}
Similar to the prior work~\cite{sun2022direct}, our optimization is mainly divided into two stages: coarse and fine. In the coarse stage, we normally train $V_{\text{den}(c)}\in\mathbb{R}^{1 \times N_x^{(c)} \times N_y^{(c)} \times N_z^{(c)}}$ with $V_{\text{rgb}(c)} \in \mathbb{R}^{3 \times N_x^{(c)} \times N_y^{(c)} \times N_z^{(c)}}$ without using Mip-VoG, as to only obtain a rough 3D geometry $V_{\text{den}(c)}$ for reducing the number of sampling points in the fine stage. 
In the fine stage, we train a density Mip-VoG $V_{\text{den}(f)} \in \mathbb{R}^{1 \times N_x^{(f)} \times N_y^{(f)} \times N_z^{(f)}}$ and a color Mip-VoG $V_{\text{rgb}(f)} \in \mathbb{R}^{7 \times N_x^{(f)} \times N_y^{(f)} \times N_z^{(f)}}$ with a tiny $\text{MLP}_\phi$ as introduced before.
We use gradient-descent to directly optimize
value in voxel grids. As the gradient of the linear interpolation used in Mip-VoG downsampling $\Downarrow_{1/2}$ is tractable, the gradient from voxel grids with different resolution can be naturally aggregated and propagated.
The loss function is the square error between the predicted pixel color and the ground truth: 
\begin{equation}
    \mathcal{L}_r = \sum\limits_{i}{\|\mathbf{C}(\mathbf{r}_i)-\widehat{\mathbf{C}}(\mathbf{r}_i)\|_2^2}.
\end{equation}

% The density of the object learned in the coarse stage will affect the performance of quering efficiency. Improving the accuracy of the density mesh will help save memory as well as speed up rendering time. In this way, we introduce an entropy loss for density voxel grid in coarse stage to encourage the density voxel grid to be sparse.
% \begin{equation}
% \begin{aligned}
%     &\mathcal{L}_c =  \\
%     &-\frac{\lambda_c}{N} \sum_{j,k}{p_{j,k}log(p_{j,k}) + (1-p_{j,k})log(1-p_{j,k})} 
% \end{aligned}
%     % \mathcal{L}_c =\\
%     % -\lambda_c\frac{1}{N} \sum_{j,k}{p_{j,k}log(p_{j,k}) + (1-p_{j,k}))log(1-p_{j,k}))}
% \end{equation}
% where $p_{j,k} = \sigma(\mathbf{r}_i(t_k))$, $\lambda_c$ is the weight for this regularization term, in our experiment, we set it as $10^{-4}$ only in coarse stage, and $j$ represents the index of pixels in the input images, $k$ is the 3D location index along a ray which is derived from $j\-th$ pixel.
%------------------------------------------------------------------------
\section{Experiments}
\label{sec:exper}
\begin{table*}[h!]
    \scriptsize
    \centering
    % \resizebox{2.\columnwidth}{!}{
    \begin{tabular}{l|cccc|cccc|cccc}
    \hline
    \multirow{2}{*}{Method} & \multicolumn{4}{c}{PSNR$\uparrow$} & \multicolumn{4}{|c}{SSIM$\uparrow$} & \multicolumn{4}{|c}{LPIPS$\downarrow$} \\
     & Full Res & 1/2 Res & 1/4 Res & 1/8 Res & Full Res & 1/2 Res & 1/4 Res & 1/8 Res & Full Res & 1/2 Res & 1/4 Res & 1/8 Res \\
     % \toprule
     \hline \hline
    \transparent{0.5}{Mip-NeRF}~\cite{barron2021mip}
    &\transparent{0.5}{32.629} &\transparent{0.5}{34.336} &\transparent{0.5}{35.471} &\transparent{0.5}{35.602} &\transparent{0.5}{0.958} &\transparent{0.5}{0.970} &\transparent{0.5}{0.979} &\transparent{0.5}{0.983} &\transparent{0.5}{0.047} &\transparent{0.5}{0.026} &\transparent{0.5}{0.017} &\transparent{0.5}{0.012} \\
    SNeRG~\cite{hedman2021baking} 
    & 27.043&28.405&30.044&28.544
    & 0.912&0.932&0.952&0.950
    & 0.100&0.067&0.047&0.049 \\
    MobileNeRF~\cite{chen2022mobilenerf} 
    & 24.115&25.127&26.633&27.930 
    & 0.868&0.885&0.913&0.938 
    & 0.141&0.112&0.078&0.050 \\
    MobileNeRF~\cite{chen2022mobilenerf} w/o SS
    &23.730& 24.425& 25.308&25.364
    & 0.861 &0.875 &0.898 & 0.910
    & 0.149 &0.128 & 0.104& 0.091\\
    Ours & \textbf{30.333} & \textbf{31.290} & \textbf{31.055} & \textbf{29.014}
    & \textbf{0.946} & \textbf{0.956} & \textbf{0.960} & \textbf{0.955}
    & \textbf{0.069} & \textbf{0.049}& \textbf{0.045} & \textbf{0.048} \\
    \hline
    \end{tabular} 
    % } 
    \caption{\textbf{Quantitative results on Multiscale-NeRF.} For comparison of models trained and evaluated on multiscale dataset. All the metrics of the scale are averaged across eight scenes. ``w/o SS'' removes supersampling from MobileNeRF.} 
    \label{tab:exp_multcam}
\end{table*}
\begin{figure*}[h!]
  \centering   \includegraphics[width=1.0\textwidth]{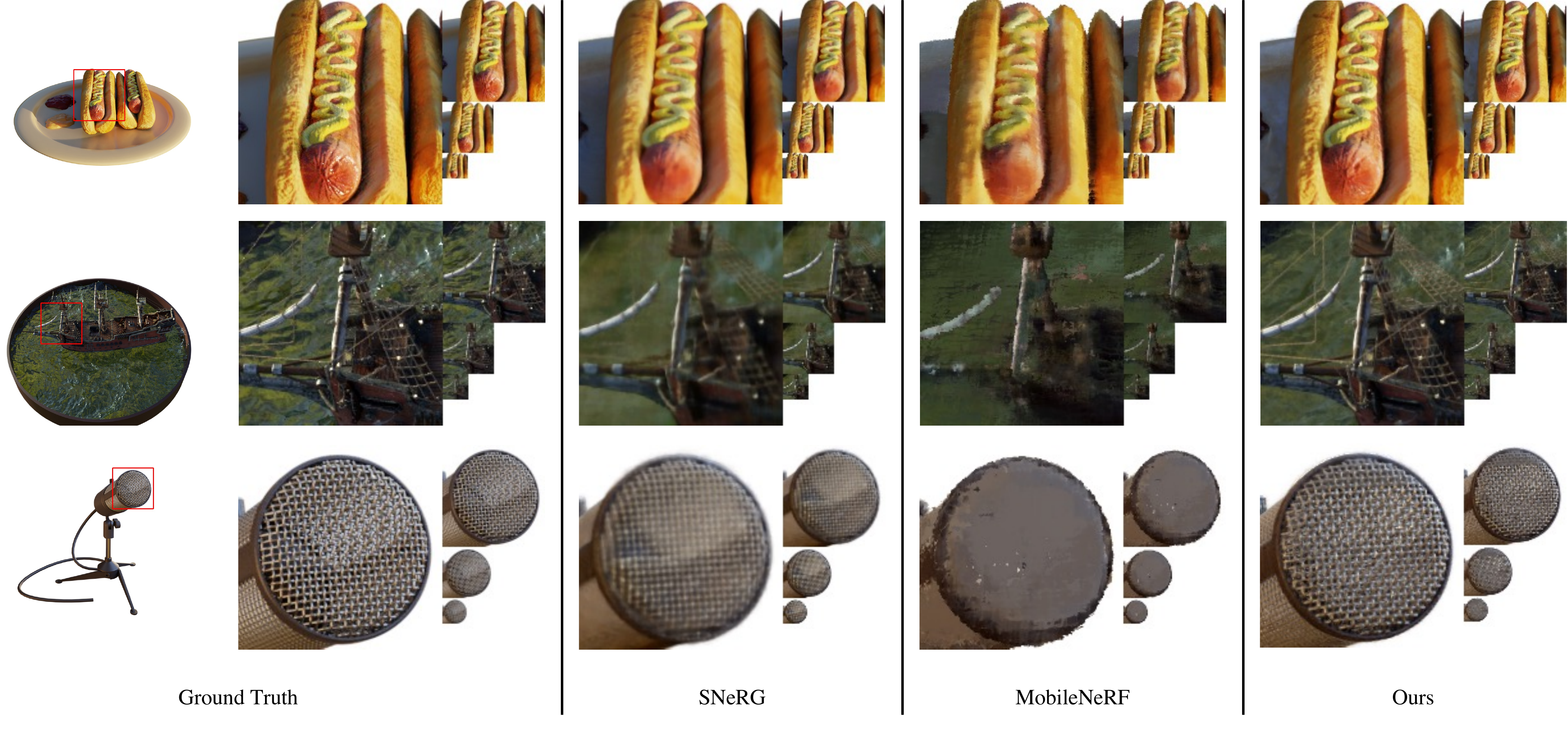}
  \caption{\textbf{Qualitative results on Multiscale-NeRF.} We demonstrate Mip-VoG rendering results compared to other baselines on the test set from three scenes, trained and evaluated on multiscale dataset. We visualize a crop region (shown in red box) on a same image at 4 different scales as an image pyramid. MobileNeRF yields over smooth results on all scales, while SNeRG lost high frequencies in high-resolution images and product aliasing in low-resolution frames. Our method surpass the baselines by a large margin as the rendering quality is significantly better.}
    \label{fig:exp_ms}
\end{figure*}
In light of our previous discussion, we primarily compare the results under the real-time rendering setting. We mainly evaluate our method on a simple multiscale synthetic dataset from Mip-NeRF~\cite{barron2021mip} designed to better validate the accuracy on multi-resolution frames. We also conduct the experiment on its single-resolution version blender dataset introduced in the original NeRF paper~\cite{mildenhall2021nerf}, in order to probe our aliasing performance of the model training on a single-scale dataset.
% , and a subset of real-world dataset Tanks and Temples~\cite{knapitsch2017tanks}. 
We report the three commonly studied error metrics: PSNR, SSIM~\cite{wang2004image}, and LPIPS~\cite{zhang2018unreasonable}, and showcase some qualitative results.
\subsection{Datasets}
\begin{enumerate}
    \item Synthetic-NeRF~\cite{mildenhall2021nerf} presented in the original NeRF paper contains eight scenes. In this single-scale dataset, each scene consists 100 training images and 200 test images with uniform 800 * 800 resolution. The model trained on this dataset can learn all the high-frequency details from the full resolution images, without being harmed by training images at multiple scales.
    
    \item Multiscale-NeRF~\cite{barron2021mip} is a straightforward conversion to Synthetic-NeRF for analyzing multiscale training and aliasing. It was generated by taking each image in Synthetic-NeRF and box down-sampling it by a factor of 2, 4, and 8 (and modifying the camera intrinsics accordingly).
    The three down-scaled images along with the original images are then combined into one single dataset. Hence this dataset contains image with four different scales for both training and test set, and the size has been quadrupled. The average evaluation metric is reported as the arithmetic mean of each error metric across all four scales. As suggested by Mip-NeRF~\cite{barron2021mip}, we adopt the Area Loss for all the methods which scale the pixel's loss by the footprint size in the full resolution images, to balance the influence between high and low resolution pixels.
    
    % \item Tanks and Temples~\cite{knapitsch2017tanks} is s single-scale real-world dataset, which has a large resolution of 1920*1080 pixels per image. Among them, it contains 5 different objects, each with manually labeled masks and known camera poses. 
    
\end{enumerate}

\subsection{Implementation Details}
In our experiments, we set the same hyperparameters for single-scale and multiscale datasets. In the coarse stage, the resolution of the voxel grid for both density and color is $(128\times 128 \times 128)$, while in the fine stage, it raises to $(512 \times 512 \times 512)$. The low pass filter is adopted as the Mean Filter with kernel size 5. We use ``shifted softplus" mentioned in Mip-NeRF \cite{barron2021mip} as the density activation. The initial values of alpha is $10^{-6}$ in the coarse training stage, and $10^{-2}$ in the fine training stage. Our tiny MLP follows the architecture used in SNeRG~\cite{hedman2021baking}. We use the Adam optimizer~\cite{kingma2014adam} to train both voxels and the deferred MLP, the learning rate are set to $4*10^{-3}$ for the deferred MLP and $1*10^{-1}$ for the voxel grids. In addition, we train 10k and 20k iterations for the coarse phase and fine phase with the batch size of 8192, respectively. For real-time web renderer, we convert Mip-VoG to the sparse voxel grid data structure~\cite{hedman2021baking} and implement our query procedure in WebGL using the THREE.js library to increase the rendering speed. In terms of a fair comparison, all the methods are trained on a 80GB A100 GPU and tested on laptop GPU. Mip-VoG enables rendering 800 $\times$ 800 images in at 52 FPS on Lenovo Legion 7 (laptop) w/ NVIDIA RTX 2070 SUPER and 71 FPS on Alienware M15R6 (laptop) w/ NVIDIA RTX 3080, with a 104 MB storage footprint.

% We report the rendering speed and storage in Tab. \ref{tab:exp_storage}. 
% \begin{table}[h!]
%     \begin{center}
%     \resizebox{0.45\textwidth}{!}{
%     \begin{tabular}{l|ccc}
%     \hline
%     Method & Storage(MB) $\downarrow$ & FPS(Dev1) $\uparrow$ & FPS(Dev2) $\uparrow$   \\
%      \hline \hline
%     SNeRG & 87 & 121 & 197 \\
%     MobileNeRF & 126 & \textbf{125} & \textbf{356}\\
%     Ours & \textbf{62} & 53 & 71 \\
%     \hline
%     \end{tabular} 
%     }
%     \end{center}
%     \caption{\textbf{Disk storage in MB and rendering speed in frames per second (FPS).} Dev1: Lenovo Legion 7 (laptop) w/ NVIDIA RTX 2070 SUPER; Dev2: Alienware M15R6 (laptop) w/ NVIDIA RTX 3080.} 
%     \label{tab:exp_storage}
% \end{table}

\subsection{Results}
\begin{figure}[h!]
    \centering
    \begin{minipage}{\linewidth}
\centerline{\includegraphics[width=\textwidth]{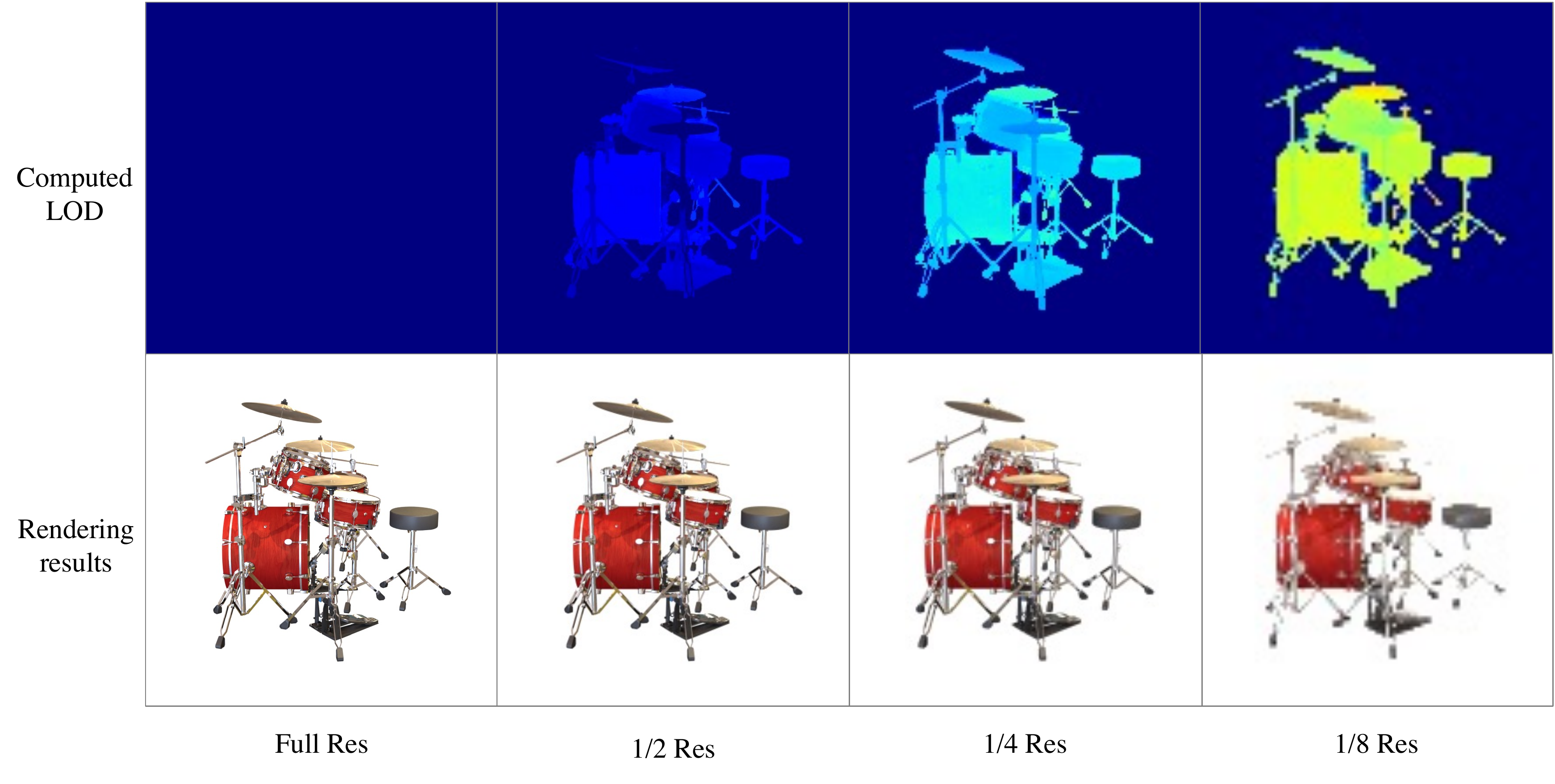}}
    \end{minipage}
    \caption{\textbf{Visualization of LOD.} We visualize the per-pixel LOD at four different scales. The value is computed using volume rendering integral of the points' LOD along the ray. We resized the low-resolution rendered images to match the dimensions of the full resolution images, enhancing visibility and facilitating direct comparison. Brighter color indicates higher values.}
    \label{fig:lod}
\end{figure}

\paragraph{Multiscale-NeRF}
The performance of Mip-VoG for this dataset can be seen in Tab.~\ref{tab:exp_multcam}. As shown in the table, our method outperforms baselines on all metrics across all scales. Note that the result is a consequence of both multi-scale training and anti-aliasing rendering, since this dataset contains multi-resolution images for both training and test set. Hence we defer the ablation into the next section. Since IPE is incompatible with current grid-based approaches that don't use PE, we only include the result of Mip-NeRF~\cite{barron2021mip} for reference purpose. Our approach, as other real-time methods, sacrifices rendering quality due to network size, resulting in lower performance compared to Mip-NeRF.
We visualize some qualitative results in Fig.~\ref{fig:exp_ms}. One can see that other real-time rendering approaches produce blurry results on high resolution images, due to the issue of multiscale training that leads the models fit on low-resolution images. In contrast, our result learns high-frequency information in the full resolution images and output low-frequencies in low resolution frames. Additionally, we visualize the computed LOD in Fig.~\ref{fig:lod}, the pixel-wise results is accumulated through the ray through volume rendering integral. 

\paragraph{Synthetic-NeRF}
\begin{table*}[h]
    \scriptsize
    \centering
    \begin{tabular}{l|cccc|cccc|cccc}
    \hline
    \multirow{2}{*}{Method} & \multicolumn{4}{c}{PSNR$\uparrow$} & \multicolumn{4}{|c}{SSIM$\uparrow$} & \multicolumn{4}{|c}{LPIPS$\downarrow$} \\
     & Full Res & 1/2 Res & 1/4 Res & 1/8 Res & Full Res & 1/2 Res & 1/4 Res & 1/8 Res & Full Res & 1/2 Res & 1/4 Res & 1/8 Res \\
     \hline \hline
    SNeRG~\cite{hedman2021baking} 
    &29.333 &30.065 &28.355 &25.373
    &0.940&0.949 & 0.946&0.924
    & 0.134& 0.091& 0.097&0.144 \\
    MobileNeRF~\cite{chen2022mobilenerf} 
    & 29.448	& \textbf{30.654}& \textbf{31.144}& \textbf{30.000}
    & 0.934&0.947 &\textbf{0.957} & \textbf{0.959}
    & 0.077 & 0.054&\textbf{0.042} &\textbf{0.037} \\
    MobileNeRF~\cite{chen2022mobilenerf} w/o SS
    &28.290& 28.447& 27.317& 25.212
    & 0.926&0.935 &0.935 & 0.917
    & 0.093& 0.077& 0.079&0.094 \\
    Ours &\textbf{30.355} & \underline{30.467} & \underline{28.766} & \underline{26.566} 
    & \textbf{0.949} & \textbf{0.956} & \underline{0.951} & \underline{0.935}
    & \textbf{0.062} & \textbf{0.050}& \underline{0.058} & \underline{0.073} \\
    \hline
    \end{tabular} 
    % } 
    \caption{\textbf{Quantitative results on Synthetic-NeRF.} Performance of models that trained on single scale Synthetic-NeRF but evaluated on Multiscale-NeRF. All the metrics of the scale are averaged across eight scenes.}
    \label{tab:exp_ssms}
\end{table*}
\begin{figure*}[h!]
  \centering
    \includegraphics[width=1.0\textwidth]{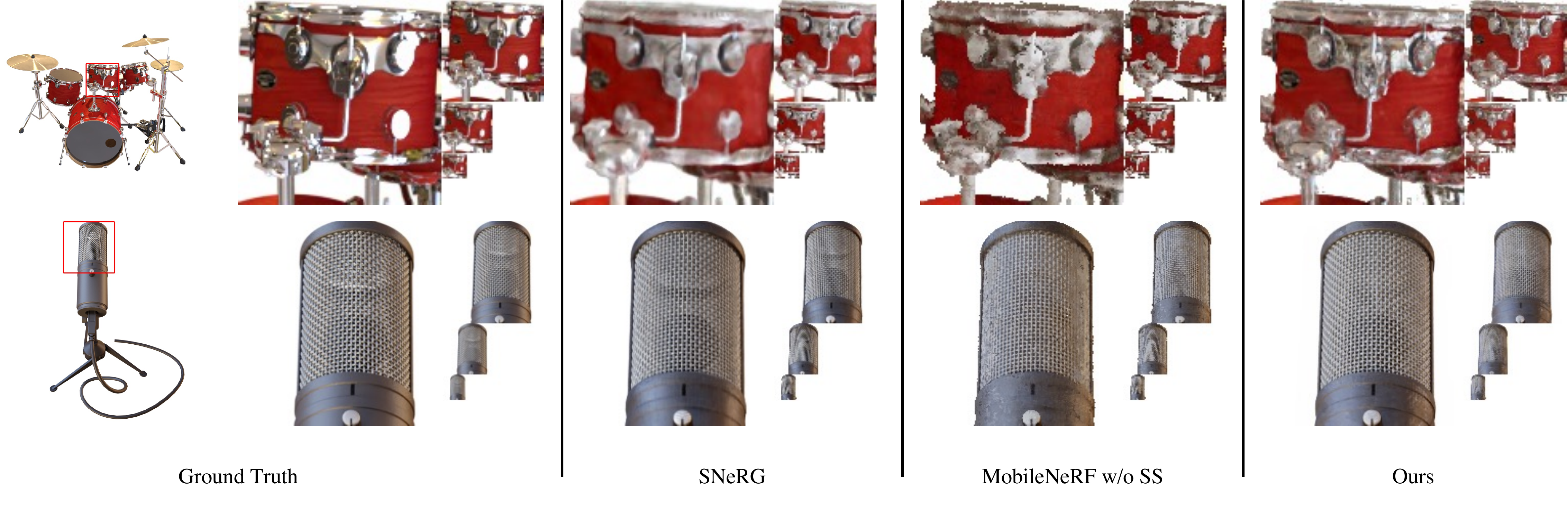}
  \caption{\textbf{Qualitative results on Synthetic-NeRF.} We demonstrate Mip-VoG rendering results compared to other baselines on the test set from two scenes, trained on single-scale dataset. One can observe that our method can reduce the aliases on the edge of drum and mic windscreen.}
    \label{fig:exp_ssms}
\end{figure*}
Since the baselines models are not compatible with multiscale training, we eliminate the factor of non-uniform training images but to examine the effectiveness on anti-aliasing rendering. For this dataset the model is trained on single scale images and evaluated on the multiscale version, since the testset Multiscale-NeRF contains all the test images in Synthetic-NeRF. This inference scenario can be seen as rendering the single scale dataset but with the distance to the viewpoint has increased by scale factors of 2, 4, and 8 (also known as minification).  In Tab.~\ref{tab:exp_ssms}, excluding the effect of multiscale training set, our method is still outperforming the baselines on all metrics on high-resolution images.
MobileNeRF~\cite{chen2022mobilenerf}, benefiting from its super-sampling technique, performs better on lower-resolution frames. 
A potential reason for the large enhancement of low-resolution renderings through super-sampling (MobileNeRF w/o SS vs. MobileNeRF) is that the final view is down-sampled from a higher-resolution rendered image, which improves accuracy and approximates the generation of ground truth low-resolution  images.
While removing super-sampling from it in the inference phase, our method outperforms the baseline models across all the scales. 
We showcase some qualitative results in Fig.~\ref{fig:exp_ssms}. One can find our method produce more low-frequency details and mitigate aliasing artifact in the low-resolution images if zooming in, while in high-resolution images our method preserve sharp high-frequency details. This results verify that our method effectively learns a multiscale representation since it improves the both multi-resolution training and anti-aliasing.

% \paragraph{Tanks and Temple} 
% The rendering quality of this single-scale real data dataset are shown in 
% Tab. \ref{tab:exp_TnT}. We exhaustly train the original implementation of SNeRG~\cite{hedman2021baking}, yet this approach always yield poor convergence results. Hence, we alternatively train a ``deferred'' DVGO~\cite{sun2022direct} as our baseline shown as SNeRG* in the table. Our methods show superior performance comparing with the other real-time methods in all metrics. However we argue that since the training and testing images come with the same scale, the margin is subtle.
% \begin{table}[h!]
% \begin{center}
% % \resizebox{0.45\textwidth}{!}{
% \begin{tabular}{l|c|c|c}
% \hline
% Method &  PSNR$\uparrow$ & SSIM$\uparrow$ & LPIPS$\downarrow$\\
% \hline\hline
%  SNeRG* & 27.020&  0.893  &  0.172 \\
%  MobileNeRF~\cite{chen2022mobilenerf} &24.559 & 0.856 & 0.183\\ 
%  Ours  & \textbf{27.043} &\textbf{0.893} & \textbf{0.172}\\ 
% \hline
% \end{tabular}
% % }
% \end{center}
% \caption{\textbf{Tanks and Temple.} For quantitative comparison of models trained and tested on real-world data. We alternatively train a deferred DVGO~\cite{sun2022direct} as the baseline (SNeRG* in the table) due to the poor convergence result of the original implementation~\cite{hedman2021baking}}
% \label{tab:exp_TnT}
% \end{table}

\subsection{Ablations}
In this section we analysis the effectiveness of the contribution of Mip-VoG to the model, and give some insights into the filtering algorithm. We perform all the experiments on Multiscale-NeRF~\cite{barron2021mip}, and report the average PSNR over the eight scenes.

\paragraph{Mipmapping}
To better examine the validity of Mip-VoG, we ablate this technique from training and testing respectively. While produce the rendering result without Mip-VoG reflects the ability of training with the images at multiple resolution, removing it from training effectively shows the improvement on anti-aliasing. As the results shown in Tab.~\ref{tab:exp_mipmapping}, training without Mip-VoG shows lower accuracy in high-resolution test images and higher quality in low-resolution frames. This result is consistent with the studies in Mip-NeRF~\cite{barron2021mip}, as the area loss would force model to ``overfit" on low-resolution training samples. While training with mip-VoG can help preserve high-frequencies rendering without Mip-VoG would produce aliasing in low-resolution frames, as the metrics of ``Ours w/o te-mip'' in lower scale is worse than the baseline. Hence, using Mip-VoG in both training and testing contributes the multiscale training and anti-aliasing rendering.
\begin{table}[!ht]
    \begin{center}
    \resizebox{0.45\textwidth}{!}{
    \begin{tabular}{l|cccc}
    \hline
    \multirow{2}{*}{Method} & \multicolumn{4}{c}{PSNR$\uparrow$} \\
     & Full Res & 1/2 Res & 1/4 Res & 1/8 Res \\
     \hline \hline
    Ours w/o tr-mip te-mip
    &  29.690 & 30.897& 30.201&27.371 \\
    Ours w/o tr-mip
    &  29.631&30.217&29.461&27.663\\
    Ours w/o te-mip
    &\textbf{30.348}&31.146&29.581&26.669 \\
    Ours & 30.333 & \textbf{31.290} & \textbf{31.055} & \textbf{29.014}\\
    \hline
    \end{tabular} 
    }
    \end{center}
    \caption{\textbf{Result of mipmapping ablation.} We conduct the experiments on the Multiscale-NeRF with respectively dropping the Mip-VoG in training and testing, denoted as ``tr-mip'' and ``te-mip'' in the table.} 
    \label{tab:exp_mipmapping}
\end{table}

\paragraph{Low-pass Filter}
One design in the sampling phase is that the voxel grids are pre-filtered by a low-pass filter, which help preserve the high frequency information in the training and abandon them in rendering. We test our model based on three type of options: no filter, gaussian filter and mean filter. We also experiment the filter with different kernel size. The results are shown in Tab.~\ref{tab:exp_filter}. Firstly, using filter yields a better performance on the rendering quality across all the scales. Secondly, Mip-VoG is robust to different filter while the superior performance is achieved when the mean filter with kernel size 5 is chosen. Finally, using too small or too large kernel size tends to slightly harm the filtering outcome, as size 5 surpass the other two for both mean filter and gaussian filter.
\begin{table}[!ht]
    \begin{center}
    \resizebox{0.45\textwidth}{!}{
    \begin{tabular}{l|cccc}
    \hline
    \multirow{2}{*}{Filter} & \multicolumn{4}{c}{PSNR$\uparrow$} \\
     & Full Res & 1/2 Res & 1/4 Res & 1/8 Res \\
     \hline \hline
    None
    & 29.995& 31.043& 30.718&28.534\\
    Mean (size 3)
    &  30.211 &31.179 & 30.997&29.005\\
    Mean (size 5)& \textbf{30.333} & \textbf{31.290} & \textbf{31.055} & \textbf{29.014}\\
    Mean (size 7)
    &  30.210 & 31.177& 30.995&29.002\\
    Gaussian (size 3)
    & 30.011& 31.068& 30.733&28.575\\
    Gaussian (size 5)
    & 30.052&31.078 & 30.759&28.683\\
    Gaussian (size 7)
    & 30.005 & 31.059& 30.731&28.505\\
    \hline
    \end{tabular} 
    }
    \end{center}
    \caption{\textbf{Low-pass filter ablation.} We demonstrate the sensitivity of the low-pass filter with respective to the filter type and the kernel size.} 
    \label{tab:exp_filter}
\end{table}
%------------------------------------------------------------------------
\section{Conclusion}
\label{sec:discuss}
In the paper, we have presented a multiscale representation for real-time anti-aliasing rendering method. We base our work on the voxel grids representation with a deferred architecture of NeRF. We proposed to use mip voxel grids, which yields the point-wise sampling from the voxel grids of different scales according to the level of detail computed by ray differentials. To generate multiple levels of Mip-VoG, we leverage the low-pass filter and interpolation filtering to downsample the original full resolution voxel grid progressively. The final scene properties of a 3D is sampled from two neighbor voxel grids using quadrilinear interpolation. Experiments show our method effectively learns a multiscale representation from the training images and provides higher accuracy in real-time anti-aliasing rendering. Our main bottleneck of rendering speed is the computation burden of Level of Detail within the web shader. We sacrifice the speed of original SNeRG formulation to provide a multiscale representation. Nonetheless, there is room for improvement through future engineering endeavors, including potential collaboration with state-of-the-art voxel-based real-time rendering engines like \cite{reiser2023merf}. While offering a pioneering multiscale representation for real-time applications, we hope our approach will be valuable to future research on multiscale training and real-time anti-aliasing rendering for neural rendering models.
%------------------------------------------------------------------------

\paragraph{\emph{\textbf{Acknowledgements}}} This research was mainly undertaken using the LIEF HPC-GPGPU Facility hosted at the University of Melbourne. This Facility was established with the assistance of LIEF Grant LE170100200.  This research was also partially supported by the Research Computing Services NCI Access scheme at the University of Melbourne. DH was supported by Melbourne Research Scholarship from the University of Melbourne.

\clearpage

{\small
\bibliographystyle{ieee_fullname}
\bibliography{egpaper}
}

\end{document}